\documentclass{llncs}
\usepackage{graphicx}
\usepackage{tikz}
\usepackage{url}

\usetikzlibrary{positioning}
\usetikzlibrary{shapes,arrows}
\usetikzlibrary{decorations.pathreplacing}
\tikzstyle{decision} = [diamond, draw=blue!10, fill=blue!20, 
    text width=3em, text badly centered, node distance=2cm, inner sep=0pt]
\tikzstyle{block} = [rectangle, draw, fill=blue!20, 
    text width=5em, text centered, rounded corners, minimum height=4em]
\tikzstyle{line} = [draw, -latex']
\tikzstyle{cloud} = [draw, ellipse,fill=red!20, node distance=3cm,
    minimum height=2em]
    
\tikzstyle{startstop} = [rectangle, rounded corners, minimum width=3cm, minimum height=0.5cm,text centered, draw=black, fill=red!30, inner sep=0]
\tikzstyle{io} = [trapezium, minimum width=2cm, minimum height=0.5cm, text centered, draw=blue!10, fill=blue!30, inner sep=0pt, trapezium stretches=true]
\tikzstyle{process} = [rectangle, minimum size=5mm, minimum height=0.4cm, text centered, text width=3cm, draw=orange!10, fill=orange!30, inner sep=0pt]
\tikzstyle{round} = [ellipse, minimum width=2cm, minimum height=0.7cm, text centered, text width=3cm, draw=blue!10, fill=blue!20, inner sep=-0.5cm]
\tikzstyle{arrow} = [thick,->,>=stealth]
    
\begin{document}
\mainmatter
\title{Deep Learning Defenses Against Adversarial Examples for Dynamic  Risk Assessment}
\titlerunning{Deep Learning Defenses for Risk Assessment}
%
\author{Xabier Echeberria-Barrio \and
Amaia Gil-Lerchundi \and Ines Goicoechea-Telleria \and
Raul Orduna-Urrutia
}
\authorrunning{X. Echeberria-Barrio et al.}
%
\institute{Vicomtech Foundation, Basque Research and Technology Alliance (BRTA), Mikeletegi 57, 20009 Donostia-San Sebastián (Spain)\\
\url{https://www.vicomtech.org}\\
\email{\{xetxeberria,agil,igoikoetxea,rorduna\}@vicomtech.org}}
\maketitle             
\begin{abstract}
Deep Neural Networks were first developed decades ago, but it was not until recently that they started being extensively used, due to their computing power requirements. Since then, they are increasingly being applied to many fields and have undergone far-reaching advancements. More importantly, they have been utilized for critical matters, such as making decisions in healthcare procedures or autonomous driving, where risk management is crucial. Any mistakes in the diagnostics or decision-making in these fields could entail grave accidents, and even death. This is preoccupying, because it has been repeatedly reported that it is straightforward to attack this type of models. Thus, these attacks must be studied to be able to assess their risk, and defenses need to be developed to make models more robust. For this work, the most widely known attack was selected (adversarial attack) and several defenses were implemented against it (i.e. adversarial training, dimensionality reduc tion and prediction similarity). The obtained outcomes make the model more robust while keeping a similar accuracy. The idea was developed using a breast cancer dataset and a VGG16 and dense neural network model, but the solutions could be applied to datasets from other areas and different convolutional and dense deep neural network models.

\end{abstract}
\section{Introduction}

Deep Neural Networks were first developed decades ago, but due to their computing power requirements, it was not until recently that they started being extensively studied and implemented in many fields that have a direct impact in our lives. Since then, they are progressively being applied and have undergone far-reaching advancements, even getting a spot in healthcare treatments or autonomous vehicles. These are very critical matters, and having a proper risk management is pressing. Some errors in the diagnostics or decision-making in these fields could potentially lead to major incidents that put people's lives at risk. This is worrisome, because it has been repeatedly reported in the literature that deep learning models are easily attacked. Thus, these attacks must be studied to be able to assess their risk and integrate it in risk analysis procedures, and defenses need to be developed to make models more robust against them.

For this work, adversarial attacks were selected, as they are ubiquitous. Thus, they are a significant parameter to take into account when analyzing and measuring risk. To work towards managing such risk, several defenses were implemented against it and compared, i.e. adversarial training, dimensionality reduction and prediction similarity. The obtained outcomes made the model more robust while preserving a similar accuracy. The idea was developed using a breast cancer dataset and a VGG16 and dense neural network (DNN) model, but it could be applied to other datasets and different convolutional neural network (CNN) and DNN deep neural network models. 

The rest of the paper is divided as follows: Section \ref{Related_work} gives an overview of the work found regarding adversarial attacks and their defenses. Section \ref{Adversarial_attacks} details the adversarial attack used, while Section \ref{Defenses} explains the proposed defenses against it. The results are given in Section \ref{Results} and Section \ref{Lessons_learned} lists the lessons that were learned.

\section{Related Work}\label{Related_work}
The adversarial attack has been widely studied in deep learning. Taking advantage of the sensitivity of the models, the attacker adds noise to a specific input sample, modifying the image imperceptibly to change the original output prediction of the sample. The first adversarial example against deep neural networks was generated using a L-BFGS method \cite{Szegedy2014IntriguingNetworks}. This discovery made researchers look into this new attack, discovering more efficient adversarial attacks, such as Fast Gradient Sing Method (FGSM) \cite{Madry2018TowardsAttacks}, Basic Iterative Method (BIM) \cite{Kurakin2019AdversarialWorld}, Projected Gradient Descent Method (PGD),  Jacobian-based Saliency Map Attack (JSMA) \cite{Papernot2016TheSettings} and DeepFool \cite{Moosavi-Dezfooli2016DeepFool:Networks}.

\begin{figure}[htp]
    \centering
    \begin{tikzpicture}[node distance=1cm]
    
        \node (input) [process] {input data};
        \node (predict) [process,below of=input] {prediction phase};
        \node (noise) [process,below of=predict] {noise obtaining
phase};
        \node (prediction) [process,below of=noise] {noise prediction
phase};
        \node (compare) [process,below of=prediction] {comparison phase};
        \node (same) [decision, below of=compare] {same?};
        \node (adversarial) [process, right of=same, xshift=2cm] {adversarial};

        \draw [arrow] (input) -- (predict);
        \draw [arrow] (predict) -- (noise);
        \draw [arrow] (noise) -- (prediction);
        \draw [arrow] (prediction) -- (compare);
        \draw [arrow] (compare) -- (same);
        \draw [arrow] (same) -- node[anchor=south] {no} (-2,-6) -- (-2,0) -- (input);
        \draw [arrow] (same) -- node[anchor=south] {yes} (adversarial);
    \end{tikzpicture}
    \caption{The procedure for obtaining an adversarial example}
    \label{fig:adversarial}
\end{figure}
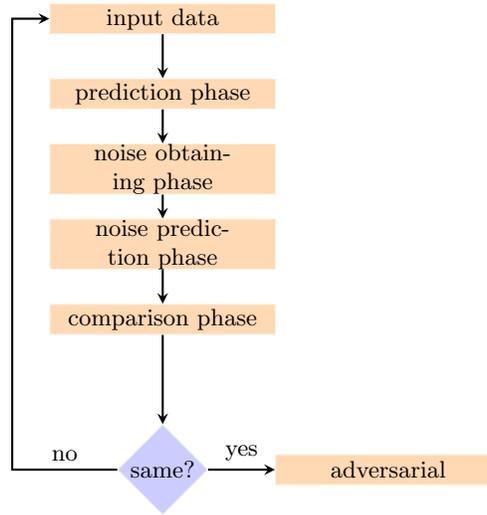

Since then, new methods have been found to avoid new adversarial attacks. Nowadays, there are two types of countermeasure strategies for adversarial examples: reactive (detecting adversarial examples once the deep neural network has been built) and proactive (making deep neural networks more robust before adversaries generate adversarial examples). This work compares three defenses: adversarial training, dimensionality reduction and prediction similarity.

\subsection{Adversarial Training (Reactive)}
This defense retrains the targeted model with the training data, once the adversarial examples have been added, so it learns to classify them correctly. This idea was introduced in \cite{Szegedy2014IntriguingNetworks}. Adversarial training is a widely used defense against adversarial attacks and it has improved over time. However, it has not achieved competitive robustness against new adversarial examples once the model is retrained \cite{Tramer2019AdversarialPerturbations,Carlini2017AdversarialMethods}.

\subsection{Dimensionality Reduction (Proactive)}

This defense can be implemented in several ways with different effectiveness in strengthening the original model, depending where new dimesionality reduction layers are inserted. However, all variants have the same idea behind: passing data through a dimensionality reduction layer (autoencoder and encoder layers, in our case) to remove as much noise as possible from the input image. Thus, the model is able to generalize, avoiding adversarial examples.
Based on reference \cite{Bhagoji2018EnhancingTransformations}, dimensionality reduction may be useful to make the targeted model more robust against adversarial examples. For the case of deep learning, CNNs and autoencoders are used to carry out the dimensionality reduction. Particularly, it is known that the autoencoders might make the model stronger against adversarial examples \cite{Sahay2019CombattingApproach,Gu2015TowardsExamples}.

\subsection{Prediction Similarity (Proactive)}
This defense adds an external layer to the model, which saves the history of parameters obtained through the input images. Adversarial attacks need several predictions of similar images to get an adversarial example. Therefore, this layer can return an adversarial probability (the likelihood that an adversarial example is being generated), after computing the similarity between the input image and previous images. If this adversarial probability is high (different from case to case) this layer could take action to avoid the adversarial attack.
There are several algorithms to compute the similarity value between two images. The most widely used metrics are the mean squared error (MSE) and peak signal to noise ratio (PSNR). However, in the last three decades, different complex metrics have been developed trying to simulate the perception of human vision by comparing two images \cite{Zhang2018TheMetric}, i.e. structure similarity metric (SSIM) \cite{Wang2004ImageSimilarity} and feature similarity metric (FSIM) \cite{Zhang2011FSIM:Lin}.  

\section{Adversarial Attack Generation} \label{Adversarial_attacks}
\begin{figure}[htp]
    \centering
    \includegraphics[width=5cm]{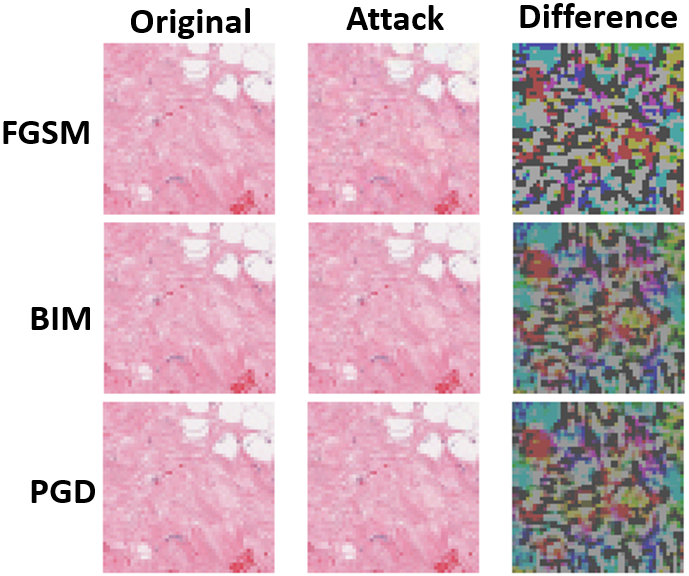}
    \caption{Original images and their adversarials. Scale was changed for clarification.}
    \label{fig:attack_types}
\end{figure}

As mentioned in Section \ref{Related_work}, the attack type chosen for this work is the adversarial attack. Particularly, we have implemented three types, namely FGSM, BIM and PGD (Fig. \ref{fig:attack_types}), through the \textit{foolbox} library\footnote{https://github.com/bethgelab/foolbox}. 

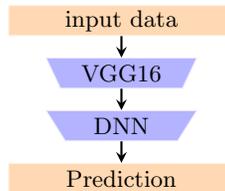
\begin{figure}[htp]
    \centering
    \begin{tikzpicture}[node distance=0.7cm]
    
        \node (input) [process] {input data};
        \node (VGG16) [io, below of=input, shape border rotate=180, minimum height=0.4cm] {VGG16};
        \node (DNN) [io, below of=VGG16, shape border rotate=180, minimum height=0.4cm] {DNN};
        \node (prediction) [process, below of=DNN] {Prediction};

        \draw [arrow] (input) -- (VGG16);
        \draw [arrow] (VGG16) -- (DNN);
        \draw [arrow] (DNN) -- (prediction);
    \end{tikzpicture}
    \caption{Our model's structure. The VGG16 could be replaced by any CNN.}
    \label{fig:model}
\end{figure}

For the experiment, a dataset of breast cancer images was used \cite{Mooney2017BreastDataset}, but could be generalized to other classification tasks. Then, we developed a model formed by a CNN (VGG16 pre-trained model) and a DNN (Fig. \ref{fig:model}), as it is widely used.

\section{Defenses to Adversarial Attacks}\label{Defenses}

Once the related work on attacks and defenses has been outlined, this section will detail the particular implementation of proposed defenses. 

\subsection{Adversarial Training}
As noted in Section \ref{Related_work}, adversarial examples are obtained and then added to the original training data. Then, the new training data is used to retrain the model (it becomes the defended model, by adversarial training). Although the new model is more robust against the added adversarial examples, it is easy to obtain new ones and stay in an inexhaustible circle of attacking and defending.

\begin{figure}[htp]
    \centering
    \begin{tikzpicture}[
    roundnode/.style={circle, draw=orange!10, fill=orange!30, thick, minimum size=2mm},
    squarednode/.style={rectangle, draw=blue!10, fill=blue!20, thick, minimum size=5mm}, node distance=0.2cm
    ]
    \node[squarednode]        (retrain)        {retrain};
    \node[squarednode]      (trained_model)     [left=of retrain]{trained model};
    \node[roundnode]        (summation)       [above=of retrain] {+};
    \node[squarednode]      (training_data)       [above=of summation] {training data};
    \node[squarednode]        (adversarial_examples)       [right=of summation] {adversarial examples};
    \node[squarednode]        (defended_model)       [below=of retrain] {defended model};
     
    \draw[->] (trained_model.east) -- (retrain.west);
    \draw[->] (training_data.south) -- (summation.north);
    \draw[->] (adversarial_examples.west) -- (summation.east);
    \draw[->] (summation.south) -- (retrain.north);
    \draw[->] (retrain.south) -- (defended_model.north);
   
    \end{tikzpicture}
    
    \caption{Standard adversarial training.}
    \label{fig:adv_training}
\end{figure}
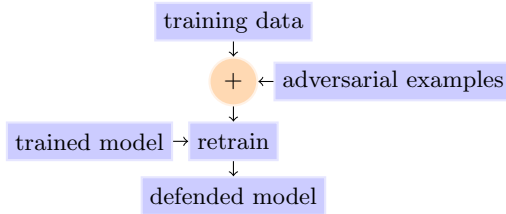

\subsection{Dimensionality Reduction}
Three variants of dimensionality reduction are covered in this subsection, which are based on the same idea, but the returned defended model is different. 

\begin{figure}[htp]
    \centering
    \begin{tikzpicture}[node distance=0.65cm]
    
        \node (input) [process] {input data};
        \node (VGG16) [io, below of=input, shape border rotate=180, minimum height=0.4cm] {VGG16};
        \node (mid1) [io, below of=VGG16, shape border rotate=180,inner sep=1pt, minimum height=0.4cm] {};
        \node (mid2) [io, below of=mid1, inner sep=1pt, yshift=0.3cm, minimum height=0.4cm] {};
        \node (DNN) [io, below of=mid2, shape border rotate=180, minimum height=0.4cm] {DNN};
        \node (prediction) [process, below of=DNN] {Prediction};

        \draw [arrow] (input) -- (VGG16);
        \draw [arrow] (VGG16) -- (mid1);
        \draw [arrow] (mid2) -- (DNN);
        \draw [arrow] (DNN) -- (prediction);
        
        \draw [decorate,decoration={brace,amplitude=10pt},xshift=-2cm,yshift=-2.4cm]
(0.5,0.5) -- (0.5,1.3) node [black,midway,xshift=-2cm] {\footnotesize middle autoencoder};
    \end{tikzpicture}
    \caption{Middle autoencoder model.}
    \label{fig:middle_autoencoder}
\end{figure}
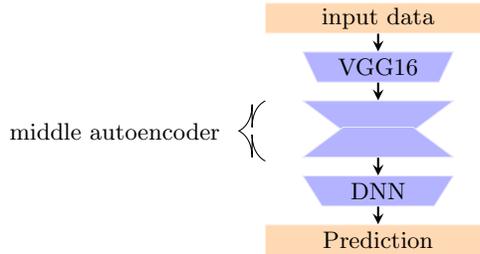

The \textbf{middle autoencoder} variant is obtained by training an autoencoder using CNN features, that is, once the outputs of data are obtained through CNN (VGG16 in our case), an autoencoder is trained using these outputs. After the autoencoder is trained, it is inserted before the DNN (Fig. \ref{fig:middle_autoencoder}). In this case, the CNN and DNN are maintained with the original structure (original weights), so they are not retrained. In short, the middle autoencoder ``cleans'' the noise of CNN's outputs before using them as DNN's input data.

\begin{figure}[htp]
    \centering
    \begin{tikzpicture}[node distance=0.7cm]
    
        \node (input) [process] {input data};
        \node (VGG16) [io, below of=input, shape border rotate=180] {VGG16};
        \node (mid1) [io, below of=VGG16, shape border rotate=180, inner sep=1pt, minimum height=0.4cm] {};
        \node (new_DNN) [io, below of=mid1, shape border rotate=180, minimum height=0.4cm, yshift=0.1cm] {New DNN};
        \node (prediction) [process, below of=new_DNN] {Prediction};

        \draw [arrow] (input) -- (VGG16);
        \draw [arrow] (VGG16) -- (mid1);
        \draw [arrow] (mid1) -- (new_DNN);
        \draw [arrow] (new_DNN) -- (prediction);
        
        \draw [decorate,decoration={brace,amplitude=5pt},xshift=-2cm,yshift=-2.15cm]
(0.5,0.5) -- (0.5,1) node [black,midway,xshift=-1.3cm] {\footnotesize encoder};
    \end{tikzpicture}
    \caption{Encoder model.}
    \label{fig:encoder}
\end{figure}
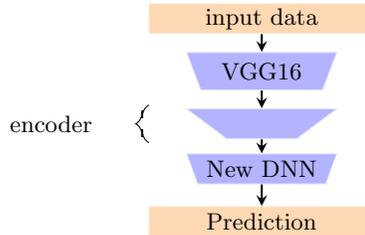

The \textbf{encoder} variant is obtained by taking the encoder part of the middle autoencoder. Then, a new model is built by inserting it between the initial CNN and a new DNN. The new DNN is trained with the encoder's output as input data and outputting the initial classes. This defense differs from the others, because the encoder trains a new DNN (Fig. \ref{fig:encoder}) so the structure of the model changes. As a summary, the encoder reduces the dimensionality of DNN's features, erasing the least important ones to avoid noise.

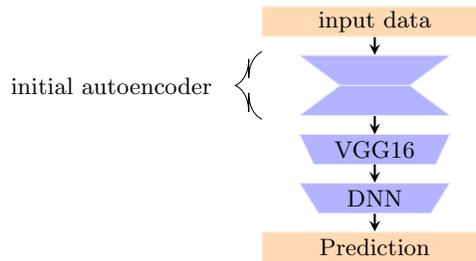
\begin{figure}[htp]
    \centering
    \begin{tikzpicture}[node distance=0.65cm]
    
        \node (input) [process] {input data};
        
        \node (mid1) [io, below of=input, shape border rotate=180, inner sep=1mm, minimum height=0.5cm, minimum height=0.4cm] {};
        \node (mid2) [io, below of=mid1, inner sep=1mm, minimum height=0.5cm, yshift=0.25cm, minimum height=0.4cm] {};
        \node (VGG16) [io, below of=mid2, shape border rotate=180, minimum width=2cm, inner sep=0, outer sep=0, minimum height=0.4cm] {VGG16};
        \node (DNN) [io, below of=VGG16, shape border rotate=180, minimum height=0.4cm] {DNN};
        \node (prediction) [process, below of=DNN] {Prediction};

        \draw [arrow] (input) -- (mid1);
        \draw [arrow] (mid2) -- (VGG16);
        \draw [arrow] (VGG16) -- (DNN);
        \draw [arrow] (DNN) -- (prediction);
        
        \draw [decorate,decoration={brace,amplitude=10pt},xshift=-2cm,yshift=-1.8cm]
(0.5,0.5) -- (0.5,1.4) node [black,midway,xshift=-2cm] {\footnotesize initial autoencoder};
    \end{tikzpicture}
    \caption{Initial autoencoder model.}
    \label{fig:initial_autoencoder}
\end{figure}

The \textbf{initial autoencoder} variant trains the autoencoder using the selected dataset and inserts it before the CNN. Both the CNN and DNN keep the original weights, since they are not retrained (Fig. \ref{fig:initial_autoencoder}). Again, the initial autoencoder ``cleans'' the image noise before making predictions with the initial model.

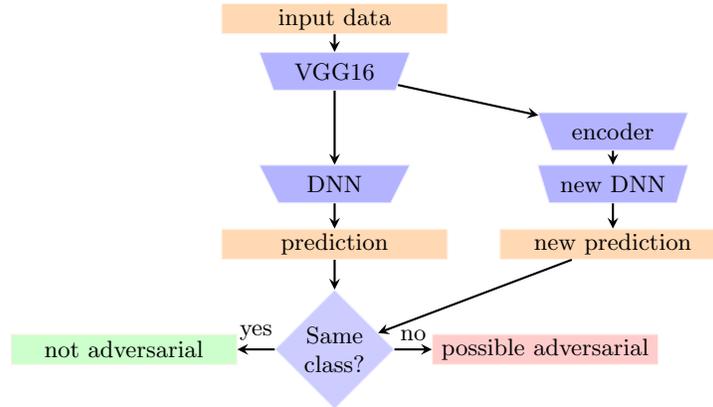
\begin{figure}[htp]
    \centering
    \begin{tikzpicture}[node distance=0.8cm]
    
        \node (input) [process] {input data};
        \node (VGG16) [io, below of=input, shape border rotate=180, yshift=0.1cm] {VGG16};
        \node (DNN) [io, below of=VGG16, shape border rotate=180, yshift=-0.7cm] {DNN};
        \node (encoder) [io, below of=VGG16, shape border rotate=180, xshift=3.7cm] {encoder};
        \node (new_DNN) [io, below of=encoder, shape border rotate=180, yshift=0.1cm] {new DNN};
        \node (prediction) [process, below of=DNN] {prediction};
        \node (new_prediction) [process, right of=prediction, xshift=2.9cm] {new prediction};
        \node (dec1) [decision, below of=prediction, yshift=0.6cm] {Same class?};
        \node (possible_adversarial) [process, right of=dec1, xshift=2cm, fill=red!20] {possible adversarial};
        \node (not_adversarial) [process, left of=dec1, xshift=-2cm, fill=green!20] {not adversarial};

        \draw [arrow] (input) -- (VGG16);
        \draw [arrow] (VGG16) -- (DNN);
        \draw [arrow] (VGG16) -- (encoder);
        \draw [arrow] (encoder) -- (new_DNN);
        \draw [arrow] (new_DNN) -- (new_prediction);
        \draw [arrow] (new_prediction) -- (dec1);
        \draw [arrow] (DNN) -- (prediction);
        \draw [arrow] (prediction) -- (dec1);
        \draw [arrow] (dec1) -- node[anchor=south] {no} (possible_adversarial);
        \draw [arrow] (dec1) -- node[anchor=south] {yes} (not_adversarial);

    \end{tikzpicture}
    \caption{External detector of adverse examples using the encoder.}
    \label{fig:dimensionality_encoder}
\end{figure}

In the case of having a trained model, the autoencoder variants could be better for their use as defense, because no parts of the original model need to be retrained. However, the encoder variant shows that a model that originally contains an encoder layer could add robustness, by using it as an adversarial detector in parallel. Therefore, we would obtain two different predictions (the original one and the one created by the defended model). In case of different predictions, the model used as defense could detect a possible adversarial example (Fig. \ref{fig:dimensionality_encoder}).

\subsection{Prediction Similarity}
As mentioned in Section \ref{Related_work}, this defense adds an external layer to the original model, which saves the history of input images used for prediction and other parameters of this action. In our case, these parameters are user, image, prediction value (the class and the probability of this class), minimum distance (to all previous images), prediction alarm (number of times the percentage of the class is smaller) and distance alarm (number of images with distance less than threshold). There are different possible actions that the output layer could take, such as blocking or predicting with a secundary model. In our case, if our layer detects something suspicious, it returns the opposite (or another) class. Thus, if the adversarial attack is detected, this action automatically avoids it, since it will return another class. This makes the adversary believe that he/she has already achieved the adversarial example, when in fact it is not. Thus, these parameters could aid decision-making outcomes and be useful for risk management measurements.

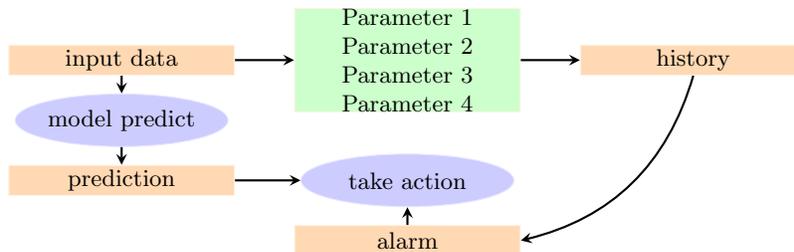
\begin{figure}[htp]
    \centering
    \begin{tikzpicture}[node distance=0.8cm]
    
        \node (input) [process] {input data};
        \node (model_predict) [round, below of=input] {model predict};
        \node (prediction) [process, below of=model_predict] {prediction};
        \node (take_action) [round, right of=prediction, xshift=3cm] {take action};
        \node (alarm) [process, below of=take_action] {alarm};
        \node (parameters) [process, right of=input, xshift=3cm, fill=green!20] {Parameter 1 \\ Parameter 2 \\ Parameter 3 \\ Parameter 4};
        \node (history) [process, right of=parameters, xshift=3cm] {history};

        \draw [arrow] (input) -- (model_predict);
        \draw [arrow] (model_predict) -- (prediction);
        \draw [arrow] (prediction) -- (take_action);
        \draw [arrow] (alarm) -- (take_action);
        \draw [arrow] (input) -- (parameters);
        \draw [arrow] (parameters) -- (history);
        \path[arrow] (history.south) edge [bend left] (alarm.east);
    \end{tikzpicture}
    \caption{Generalization of the prediction similarity defense.}
    \label{fig:similar_prediction}
\end{figure}

\section{Results}\label{Results}
This section will detail the results obtained with each defense through two types of adversarial examples: the initial model's adversarial examples and the defended model's adversarial examples (new adversarials).

\textbf{Initial adversarial examples: }All three defenses have been tried on this case. Each defense was tested to calculate how many initial adversarial examples are no longer misclassified. Adversarial training is the best option of the three to avoid initial adversarial examples (Tab. \ref{tab1}), as the model is retrained with the initial adversarial examples directly. Dimensionality reduction defends against this type of attack, while the prediction similarity does not, since the process of generating initial adversarials has already been carried out. It merely detects when an adversarial attack attempt is happening, which is useful as a parameter for risk assessment.

\begin{table}
\centering
\caption{Percentage of known adversarial examples that are no longer adversarials and how our defended models behave with new adversarial examples. Computed 3 times and averaged.}\label{tab1}
\begin{tabular}{|l|c|}
\hline
{\bfseries Defendes} & {\bfseries Initial adversarials}\\
\hline
Adversarial training &  92.0\%\\
Middle autoencoder & 60.4\%\\
Encoder &  64.3\%\\
Initial autoencoder & 70.5\%\\
\hline

\end{tabular}
\end{table}

\begin{figure}[htp]
    \centering
    \includegraphics[width=10cm]{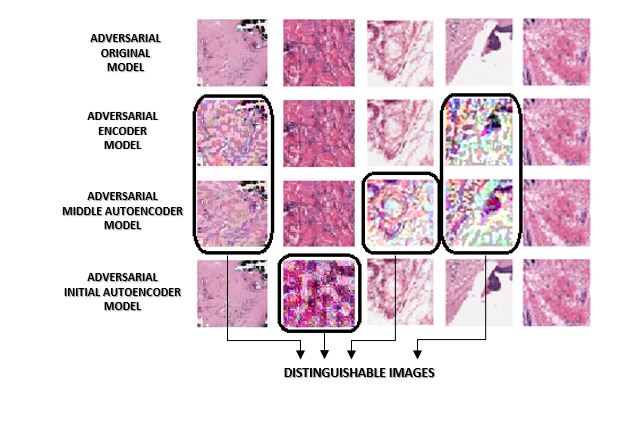}
    \caption{Image results of the different dimensionality reduction defenses.}
    \label{fig:dimensionality_results}
\end{figure}

\textbf{New adversarial examples: }Once the defenses had been tested with initial adversarial examples, new ones were generated to attack the defended model. In this case, the adversarial training is not at all robust, as it is easy to get new adversarial examples of the defended model. However, dimensionality reduction is more robust in this case, as is visible in (Fig. \ref{fig:dimensionality_results}), since the new adversarials become distinguishable for the human-eye. Finally, the similarity of the predictions is the one that detects the greatest number of generation processes of these new adversarials (\textbf{99.5\% detection success}). The difficulty of this defense is selecting adequate parameters and thresholds, and these can change depending on the dataset, the adversarial attack and the chosen metric. In our case, it has been implemented with the parameters from Section \ref{Defenses} and the SSIM metric.

As far as we know, this is the first time that this type of dimensionality reduction and prediction similarity have been proposed as defenses for adversarial examples (which are the hardest to avoid). In addition, concretely for the dynamic risk analysis case, the prediction similarity defense is useful because it is an attack detection approach that could give meaningful insights for calculating risk levels.

\section{Lessons Learned}\label{Lessons_learned}

For this work, a widely known attack was selected (adversarial attack) and several defenses were implemented against it (i.e. adversarial training (that was used for comparison), dimensionality reduction and prediction similarity (the proposed two new defends). The obtained outcomes make the model more robust while maintaining a similar accuracy. The idea was developed using a breast cancer dataset and a VGG16 model, but the solutions could be applied to datasets from other areas and different CNN and DNN models. The highlights from different defenses studied in this paper are the following:
\begin{itemize}
    \item \textbf{Adversarial training}: it is not helpful because new adversarials can be generated, so it becomes an endless circle.
    \item \textbf{Dimensionality reduction}: it works when looking for new adversarials, because the generated noise becomes perceptible for the human eye. Also, it keeps the accuracy stable while making the model more robust.
    \item \textbf{Prediction similarity}: it has the advantage of not having to modify the model and it could be a useful input for risk assessment, as it detects when an attack is being carried out with a high accuracy.
\end{itemize}

In the future, these defenses could be applied to other types of machine learning, such as reinforcement learning, and they could be integrated in risk assessment measurements.

\section{Acknowledgements}
 
This work is funded under the SPARTA project, which has received funding
from the European Union Horizon 2020 research and innovation
programme under grant agreement No 830892.
%
%
%
%

\bibliography{bibliography} 
\bibliographystyle{splncs04}
\end{document}